\documentclass[10pt,twocolumn,letterpaper]{article}

\usepackage{iccv}              
\usepackage[accsupp]{axessibility}     

\usepackage[T1]{fontenc}
\definecolor{iccvblue}{rgb}{0.21,0.49,0.74}
\usepackage[pagebackref,breaklinks,colorlinks,allcolors=iccvblue]{hyperref}

\usepackage{algorithm}
\usepackage{algorithmic}
\usepackage{amsmath}
\usepackage{graphics}
\usepackage{epsfig}

\usepackage{natbib}

\def\paperID{11925} 
\def\confName{ICCV}
\def\confYear{2025}

\title{NAPPure: Adversarial Purification for Robust Image Classification under Non-Additive Perturbations}

\author{Junjie Nan\textsuperscript{1,2}, Jianing Li\textsuperscript{1*}, Wei Chen\textsuperscript{1,2}, Mingkun Zhang\textsuperscript{1,2}, Xueqi Cheng\textsuperscript{1,2} \\
\textsuperscript{1}State Key Laboratory of AI Safety, Institute of Computing Technology, \\Chinese Academy of Sciences, Beijing 100190, China\\
\textsuperscript{2}University of Chinese Academy of Sciences, Beijing, 101408, China\\
{\tt\small \{{nanjunjie23s, lijianing, chenwei2022, zhangmingkun20z, cxq}\}@ict.ac.cn}
}

\begin{document}
\maketitle
\begin{abstract}
Adversarial purification has achieved great success in combating adversarial image perturbations, which are usually assumed to be additive. However, non-additive adversarial perturbations such as blur, occlusion, and distortion are also common in the real world. Under such perturbations, existing adversarial purification methods are much less effective since they are designed to fit the additive nature. In this paper, we propose an extended adversarial purification framework named NAPPure, which can further handle non-additive perturbations. Specifically, we first establish the generation process of an adversarial image, and then disentangle the underlying clean image and perturbation parameters through likelihood maximization. Experiments on GTSRB and CIFAR-10 datasets show that NAPPure significantly boosts the robustness of image classification models against non-additive perturbations.
\end{abstract}    
\section{Introduction}
\label{sec:intro}
\renewcommand\thefootnote{}
\footnote{* Corresponding author: lijianing@ict.ac.cn}
Neural networks are vulnerable to adversarial attacks: in image classification, an imperceptible perturbation added to the input image can lead to misclassification with high confidence  \cite{szegedy2013intriguing} .  This phenomenon has raised concerns about the reliability of deep learning models in risk-sensitive applications such as autonomous driving, intelligent security system, and smart healthcare. Adversarial perturbations are usually assumed to be additive and imperceptible, which is implemented by directly adding a scale-limited and carefully-crafted noise image to the original image \cite{goodfellow2014explaining,kurakin2016adversarial}. However, non-additive perturbations such as blur, occlusion, and distortion are common in real applications, and may even be physically easier to exploit by attackers \cite{brown2017adversarial,eykholt2018robust,liu2018dpatch,kanbak2018geometric,xu2020adversarial,wang2023adversarial}. For example, blur and distortion can be implemented by attaching optical films, and occlusion by attaching printed patches. Recent researches have shown that, non-additive perturbations are also effective for adversarial attacks, leading to significantly decreased performance of image classification systems \cite{guo2020abba,rao2020adversarial,zhang2020generalizing}.
\begin{figure}[t]
\begin{center}
\centerline{\includegraphics[width=0.7\columnwidth]{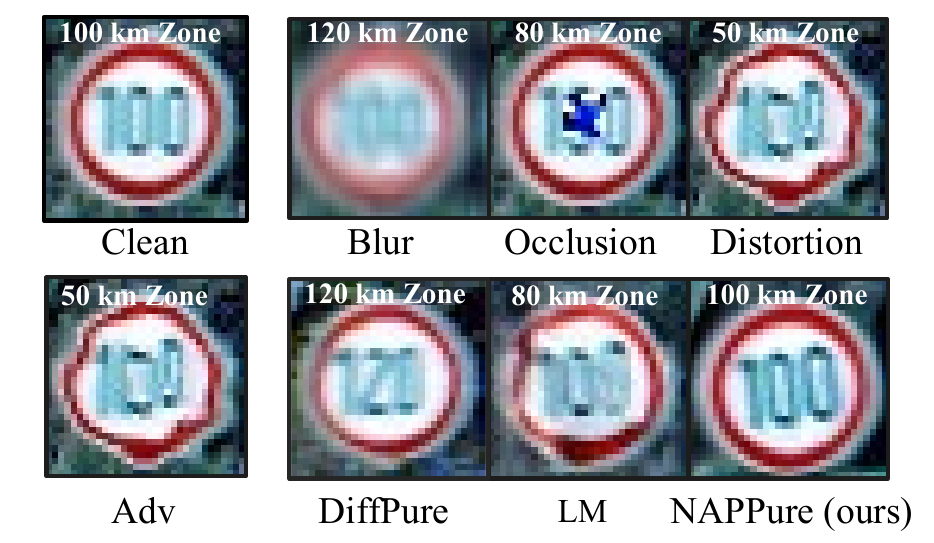}}
\caption{Upper: typical adversarial attacks with non-additive perturbations. Lower: adversarial image (Adv) and corresponding purification results. White texts are classified labels.}
\label{attack_case}
\end{center}
\vskip -0.4in
\end{figure}

Adversarial purification has been proved to be an effective approach against additive adversarial attacks \cite{nie2022diffusion,chen2023robust,zhang2024classifier}. The idea is to remove or reduce the adversarial perturbations by preprocessing the input data before feeding it to the classifier, thereby enhancing its robustness. Our observations, as shown in Fig.~\ref{attack_case}, indicate that existing adversarial purification methods (DiffPure \cite{nie2022diffusion} and LM \cite{chen2023robust}) are significantly less effective when dealing with non-additive perturbations. The reason may be that, non-additively perturbed images cannot be easily modeled by additive perturbations under restricted scales. Though any transformations can be viewed as an additive one by treating the difference as the additive perturbation, the scale (such as $l_2$ norm) of perturbation may be large enough to accommodate multiple feasible solutions from different classes. In other words, if additive adversarial purification method is forcibly applied to handle non-additive perturbations, adverse outcomes such as semantic drift may be inevitable, leading to incorrect classification result.

In this paper, we aim to extend the application scope of adversarial purification to general types of perturbations, especially for non-additive ones. Specifically, we propose the \textit{Non-Additive Perturbation Purification (NAPPure)} framework (Fig.~\ref{fig:main}), which is able to tackle non-additive adversarial attacks, where the general type of perturbation is known in advance, but the decisive parameters remain unknown.  The key idea is, model the generation process of perturbed image as a transformation from clean image and perturbation parameters, and then optimize both inputs through likelihood maximization with a pretrained diffusion model, i.e., search for the most possible combination that can recover the perturbed image. In this way, the clean image and perturbation parameters can be disentangled, then the clean image can be fed into downstream tasks.

Our NAPPure framework has the following beneficial features: when the perturbation type is additive, NAPPure naturally degenerates into traditional adversarial purification method, thus is a compatible extension; composite non-additive transformations can be obtained by aggregating multiple simple transformations, 
and each component can be applied in a plug-and-play manner; prior knowledge about perturbation parameters can be naturally integrated by introducing an individual loss term.

We implement NAPPure for 3 typical non-additive perturbations: convolution based transformation for blur \cite{guo2020abba}, patch based transformation for occlusion \cite{rao2020adversarial}, and flow-field based transformation for distortion \cite{zhang2020generalizing}. These transformations as well as their combinations are representative for a wide range of non-additive perturbations commonly encountered in real-world scenarios. On GTSRB dataset, NAPPure achieves an average robust accuracy of 70.8\% against non-additive perturbations. In contrast, traditional adversarial purification method achieves only 43.2\%, while standard adversarial training achieves 33.8\%. Such results highlight the superiority of NAPPure in terms of robustness. 
We conclude our contributions as follows: 
\begin{enumerate}[leftmargin=*, align=left]
    \item We propose NAPPure, a novel adversarial purification method against adversarial attacks with general perturbation types, especially for non-additive ones.
    \item We provide implementations for 3 typical types of non-additive perturbations under NAPPure framework, offering guiding examples for the design of further types.
    \item We demonstrate through experiment that, NAPPure is effective against adversarial attacks with non-additive perturbations, while traditional adversarial purification methods are not. 
\end{enumerate}
\section{Related Work}
\label{sec:Related Work}

\textbf{Adversarial attack.} 
The original purpose of developing adversarial attacks is detecting the vulnerability of neural networks, typically through introducing imperceptible perturbations. Main stream attack methods all focuses on additive perturbations, such as FGSM \cite{goodfellow2014explaining}, BIM \cite{kurakin2016adversarial}, PGD \cite{mkadry2017towards}, and SquareAttack \cite{andriushchenko2020square}. Under small perturbations, the semantics of an image is assumed to be unchanged, while the magnitude of an additive perturbations is defined by its $l_2$ or $l_\infty$ norm. However, the semantics of an image can be unchanged under non-additive perturbations as well, exhibiting different types of vulnerability. Such attacks have also been explored by researchers, including adversarial blur attacks \cite{guo2020abba,guo2021learning}, adversarial patches \cite{brown2017adversarial,liu2018dpatch,eykholt2018robust,ren2020adversarial,mi2023adversarial}, and geometric attacks \cite{kanbak2018geometric,zhang2020generalizing}. Defending against both additive and non-additive perturbations are important, since all such attacks are feasible threats for the safety of downstream applications. Therefore, our work focuses on defending against general types of attacks.

\textbf{Adversarial defense.} 
There are two prominent approaches for adversarial defense: adversarial training and adversarial purification. Adversarial training involves training the model with adaptively generated adversarial examples to improve model robustness \cite{mkadry2017towards,dong2023enemy,zhao2024adversarial}. While being effective and can be naturally extended to any types of perturbations, adversarial training is less effective when facing with unseen attacks \cite{chen2023robust,zhang2024causaldiff}, and model retraining is inevitable for remedy each time a new attack type emerges. Adversarial purification, as another approach, aims to eliminate adversarial perturbations from the input image before it is fed into the downstream classifier \cite{chakraborty2021survey,kalaria2022towards,nie2022diffusion,chen2023robust,song2024mimicdiffusion}. Though exhibit strong defense abilities, existing adversarial purification methods are mainly evaluated under additive perturbations, and their performance under non-additive perturbations have not been fully examined.
Some studies have attempted to address adversarial attacks with non-additive perturbations. \citet{kanbak2018geometric} try to detect and defend geometric attacks through geometric transformation invariance techniques. \citet{guo2021learning} proposes a method of learning to adversarially blur in the context of visual object tracking, showing the potential of blur as a non-additive perturbation in adversarial scenarios. Meanwhile, \citet{yu2022defending,chen2023defending,huang2024patchbreaker} focus on combating patch attacks by analyzing patch characteristics and applying patch filtering methods. Such works are limited to specific types of non-additive perturbations and cannot be applied to other types. In this paper, we aim to proposes a general adversarial purification framework for handling various non-additive attacks.

\textbf{Image restoration.} 
Image restoration is a task related to adversarial purification, which focuses on recovering the image from common corruptions without considering an adversary. Typical image restoration methods include techniques like de-blur \cite{yan2017image,chu2022improving}, inpainting for occlusion removal \cite{yeh2017semantic,tang2024realfill}, and geometric transformation correction \cite{zhang2023deformation}. Such methods are usually specifically designed for corresponding perturbation types, thus cannot tackle general types. The work of \citet{zhu2023denoising} tries to propose a framework for denoising general perturbations. However, it is unable to handle perturbations with unknown parameters, which is common in real applications. While image restoration tasks generally consider naturally-happened corruptions, adversarial attacks can be viewed as worst-case corruptions, and we focus on this harder task.

\section{Preliminary}

\begin{figure*}[t]
    \centering
    \includegraphics[width=1.4\columnwidth]{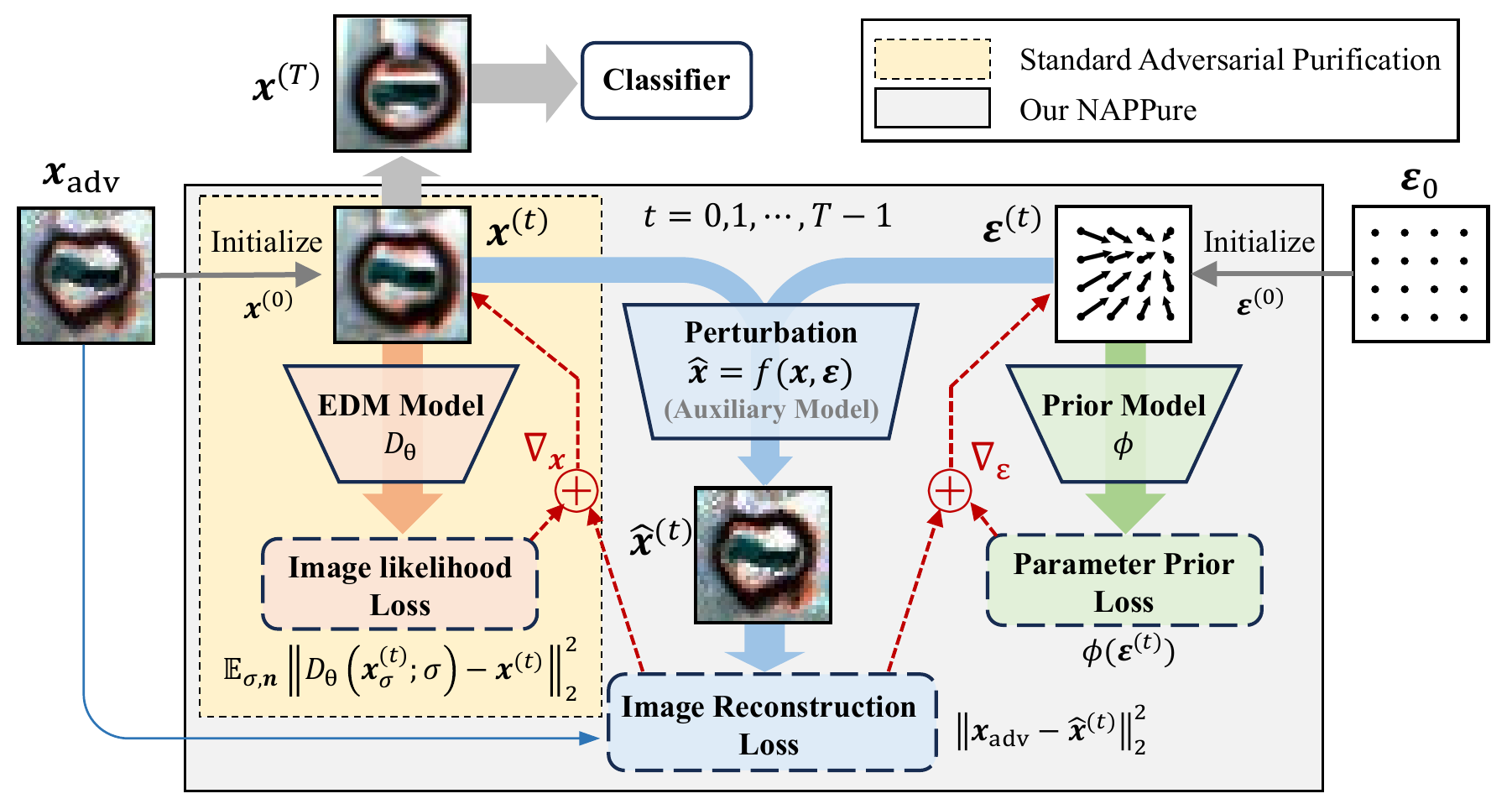}
    \caption{The main procedure of our NAPPure algorithm (images and perturbation parameters are illustrative examples under the flow-field based transformation). NAPPure purifies the adversarial image by jointly updating the underlying clean image and the perturbation parameter, before feeding the final image to downstream classifier. }
    \label{fig:main}
\end{figure*}

In this section, we will introduce our problem setting and discuss the methodology of typical adversarial purification methods.

\subsection{Problem Setting}\label{sec:setting}
We focus on the adversarial attack problem in image classification task with non-additive perturbations. Given an RGB image $\mathbf{x}\in [0, 1]^{3\times h\times w}$ as well as its ground truth label $y\in [C]=\{1,2,\cdots,C\}$, a classifier $c:[0, 1]^{3hw} \to \mathbb{R}^C$ outputting logits, then an adversarial perturbation with parameters $\mathbf{\varepsilon}\in \Omega \subseteq \mathbb{R}^d$ is constructed by the maximizing the classification loss:
\begin{equation}
\underset{\mathbf{\varepsilon} \in \Omega}{\text{max}} \ \mathcal{L}(c(f(\mathbf{x},\mathbf{\varepsilon})), y)
\end{equation}
where $f:[0, 1]^{3hw}\times \Omega \to [0, 1]^{3hw}$ is the transformation function, $\mathcal{L}:\mathbb{R}^C\times [C]\to \mathbb{R}$ is the classification loss function such as cross-entropy loss, $\Omega$ is the parameter domain which keeps the semantics of $\mathbf{x}$ unchanged after perturbation. Setting transformation $f$ to additive, i.e. $f(\mathbf{x},\mathbf{\varepsilon})=\mathbf{x}+\mathbf{\varepsilon}$, will result in the traditional adversarial attack. However in this work, we allow $f$ to be non-additive and consider the following typical transformations:

\begin{itemize}[leftmargin=*, align=left]
\item \textbf{Blur} (Convolution based transformation) \cite{guo2020abba}: $f_\mathrm{blur}(\mathbf{x}, \mathbf{\varepsilon})=\mathbf{x} * \mathbf{\varepsilon}$, where $*$ is the convolution operation and $\mathbf{\varepsilon}\in\mathbb{R}^{k\times k}$ is the kernel with size $k$.
\item \textbf{Occlusion} (Patch based transformation) \cite{rao2020adversarial}: $f_\mathrm{occl}(\mathbf{x}, \mathbf{\varepsilon})=\mathbf{x}\cdot (1-\mathbf{m})+\mathbf{p}\cdot \mathbf{m}$, where $\mathbf{\varepsilon}=(\mathbf{p},a,b,s)$, $\mathbf{p}\in\mathbb{R}^{3hw}$ is the patch pattern, $a\in [h],b\in [w]$ are the coordinates of the patch, $s\in\{0,\cdots,s_\mathrm{max}\}$ is the size of the patch. $\mathbf{m}\in\{0,1\}^{hw}$ is a binary mask with $m_{j,k}=1$ iff. $a\leq j< a+s,b\leq k< b+s$. 

\item\textbf{Distortion} (Flow-field based transformation) \cite{zhang2020generalizing}: $f_\mathrm{dist}(\mathbf{x}, \mathbf{\varepsilon})=\mathbf{x}'$, where $\mathbf{\varepsilon}\in[0,1]^{2hw}$ is a 2-dimensional flow field causing position shift of pixels. Each pixel $x'_{:,j,k}$ in $\mathbf{x}'$ is calculated by a weighted average of original pixels around the target position, i.e., $x'_{:,j,k}=\sum_{(j',k')\in N(j+\varepsilon_{0,j,k},k+\varepsilon_{1,j,k})} x_{:,j',k'}\cdot(1 - |\varepsilon_{0,j,k} - \lfloor \varepsilon_{0,j,k}\rfloor|)(1 - |\varepsilon_{1,j,k} - \lfloor \varepsilon_{1,j,k}\rfloor|)$, where $N(j,k)=\{(\lfloor j\rfloor,\lfloor k\rfloor),(\lfloor j+1\rfloor,\lfloor k\rfloor),(\lfloor j\rfloor,\lfloor k+1\rfloor),(\lfloor j+1\rfloor,\lfloor k+1\rfloor)\}$ denotes the four neighboring grid vertices of the shifted coordinate $(j,k)$.


\end{itemize}

The goal of adversarial purification is to build an inverse process $g:[0, 1]^{3hw} \to [0, 1]^{3hw}$, converting the perturbed image $\mathbf{x}_{\mathrm{adv}}=f(\mathbf{x},\mathbf{\varepsilon})$ into a purified one $\tilde{\mathbf{x}}=g(\mathbf{x}_{\mathrm{adv}})$, so that $\tilde{\mathbf{x}}$ can be correctly classified as $y$ by $c$. The transformation function $f$ is assumed to be known in advance, but the corresponding parameter $\mathbf{\varepsilon}$ remains unknown.

\subsection{Adversarial Purification}

Typical adversarial purification methods adopt the idea that perturbed images are bad samples that deviate from data manifold. To drive such samples back to the manifold, generation models such as diffusion models that captures the probability distribution of data are usually incorporated. The purification is then achieved by updating the perturbed image according to probability density ascent, or in other words, maximizing the likelihood of the image under the generation model. Therefore, adversarial purification include 2 key components: estimation of the probability density (or the gradient), and updating rule of the image.

\textbf{Density estimation.} Though generation models such as GANs \cite{goodfellow2014generative} or VAEs \cite{kingma2013auto} are ever applied \cite{samangouei2018defense, li2019defense}, mainstream adversarial purification methods adopt diffusion models \cite{ho2020denoising, song2020score} for estimation of the image distribution, due to their powerful ability on generating high-quality images. For example, continuous-time diffusion models use time-dependent score function $s_{\mathbf{\theta}}(\mathbf{x},t)$ to approximate the gradient $\nabla_{\mathbf{x}} \log p_t(\mathbf{x})$ through denoising score matching \cite{song2020score}, where $p_t(\mathbf{x})$ is the diffused distribution at time step $t$ and $p_0(\mathbf{x})=p(\mathbf{x})$. 

\textbf{Updating rule}. A straightforward approach is to directly update the image towards higher likelihood. For example, \citet{chen2023robust}
adopt an objective which maximize the evidence lower bound (ELBO) of $\log p_{\mathbf{\theta}}(\mathbf{x})$.
Another approach is utilizing the backward process of the diffusion model, either by sampling new images from scratch with guidance from input image \cite{wang2022guided, song2024mimicdiffusion}, or run forward process to time $t=t^*$ followed by backward process back to $t=0$ \cite{nie2022diffusion}. The backward process mainly improves image likelihood according to the score function $s_{\mathbf{\theta}}(\mathbf{x},t)$.

Aforementioned adversarial purification methods are motivated from the denoising ability of diffusion models and did not take into consideration of the generation mechanism of an adversarial example. We name them \textit{standard adversarial purification} in this paper, and will show that they are specific solutions for additive perturbations in Sec.~\ref{sec:algorithm}.

\section{Method}
In this section, we introduce our \textit{Non-additive Perturbation Purification} (NAPPure) framework.

\subsection{Overall Objective}

The main idea of our proposed framework is to disentangle the underlying clean image $\mathbf{x}$ and the perturbation parameter $\mathbf{\varepsilon}$ from a given adversarial image $\mathbf{x}_{\mathrm{adv}}=f(\mathbf{x},\mathbf{\varepsilon})$. Following the likelihood maximization paradigm, our goal is to find the best $\mathbf{x}^*$ and $\mathbf{\varepsilon}^*$ with highest log-likelihood $\log p(\mathbf{x},\mathbf{\varepsilon}|\mathbf{x}_\mathrm{adv})$, which can be decomposed according to the generation mechanism of $\mathbf{x}_\mathrm{adv}$:
\begin{equation}
\log p(\mathbf{x},\mathbf{\varepsilon}|\mathbf{x}_\mathrm{adv})=\log \frac{p(\mathbf{x})\cdot p(\mathbf{\varepsilon})\cdot p(\mathbf{x}_\mathrm{adv}|\mathbf{x},\mathbf{\varepsilon})}{p(\mathbf{x}_\mathrm{adv})}.
\end{equation}
Since the denominator is irrelevant to the optimization, the objective is equivalent to the following:
\begin{equation}\label{eqn:objective}
\mathbf{x}^*,\mathbf{\varepsilon}^*=\underset{\mathbf{x},\mathbf{\varepsilon}}{\text{max}} \log p(\mathbf{x}) +\log p(\mathbf{\varepsilon}) + \log p(\mathbf{x}_\mathrm{adv}|\mathbf{x},\mathbf{\varepsilon}).
\end{equation}
In this objective,
\begin{itemize}[leftmargin=*, align=left]
    \item The \textit{image likelihood} term $\log p(\mathbf{x})$ pushes the perturbed image towards areas with higher probability density, eliminating potential perturbations, which is the essential component of adversarial purification. This term can be captured by a diffusion model.
    \item The \textit{perturbation prior} term $\log p(\mathbf{\varepsilon})$ generally penalize perturbations with overly high magnitude, inhibiting potential search for unreasonable perturbations. This term can be chosen according to human knowledge.
    \item The \textit{image reconstruction} term $\log p(\mathbf{x}_\mathrm{adv}|\mathbf{x},\mathbf{\varepsilon})$ restricts the solutions to be valid under the generation process of $\mathbf{x}_{\mathrm{adv}}$, avoiding solutions that cause unintended semantic drift. This term can be modeled according to the known transformation $\mathbf{x}_{\mathrm{adv}}=f(\mathbf{x},\mathbf{\varepsilon})$.
\end{itemize}

\subsection{Optimization Loss}
We aim to solve the above optimization problem through gradient descent. Each term in the above objective corresponds to one loss term. We design them as follows.

\textbf{Image Likelihood $\log p(\mathbf{x})$.} We adopt EDM \cite{karras2022elucidating}, a strong diffusion model, to model the data distribution. We then follow the ELBO approximation approach \cite{chen2023robust} and use the following term to approximate the likelihood:
\begin{eqnarray}
-\mathbb{E}_{\sigma,\mathbf{n}}\left[\lambda(\sigma)\left\|D_{\mathbf{\theta}}(\mathbf{x}_\sigma;\sigma)-\mathbf{x}\right\|_{2}^{2}\right],
\end{eqnarray}
where $\mathbf{x}_\sigma=\mathbf{x}+\sigma\cdot\mathbf{n}$, $\sigma\sim p_\mathrm{data}(\sigma)$ is the noise level, $\mathbf{n}$ is the random Gaussian noise, $\lambda(\sigma)$ is the loss weight. $D_{\mathbf{\theta}}(\mathbf{x}; \sigma)=c_\mathrm{skip}(\sigma)\cdot\mathbf{x}+c_\mathrm{out}(\sigma)\cdot F_{\mathbf{\theta}}(c_\mathrm{in}(\sigma)\cdot\mathbf{x};c_\mathrm{noise}(\sigma))$
is the denoiser, where $F_{\mathbf{\theta}}$ is the neural network in EDM, and $c_\mathrm{skip}(\sigma),c_\mathrm{out}(\sigma),c_\mathrm{in}(\sigma),c_\mathrm{noise}(\sigma)$ are scaling weights. In practice, we set $\lambda(\sigma)=1$ for simplicity.

\textbf{Perturbation prior $\log p(\mathbf{\varepsilon})$.} We use an energy-based model to represent this term:
\begin{eqnarray}
\log p(\mathbf{\varepsilon})=\log \frac{1}{Z}e^{-\phi(\mathbf{\varepsilon})}=-\phi(\mathbf{\varepsilon})-\log Z,
\end{eqnarray}
where $\phi(\mathbf{\varepsilon})$ is a potential function, such as $l_2$ norm; $Z=\int_\Omega e^{-\phi(\mathbf{\varepsilon})}\mathrm{d}\mathbf{\varepsilon}$ is the partition function which is irrelevant to $\mathbf{\varepsilon}$. So that the final term for optimization is $-\phi(\mathbf{\varepsilon})$. In general, the potential function should take its minimal value at the identity element $\mathbf{\varepsilon}_0$ of the perturbation function, such that $f(\mathbf{x},\mathbf{\varepsilon}_0)=\mathbf{x}$. In this way, when taking a clean image as input, the purification process will tend to keep it unchanged. Note that the identity element may not be unique.

\textbf{Image reconstruction $\log p(\mathbf{x}_\mathrm{adv}|\mathbf{x},\mathbf{\varepsilon})$.} The ideal solution is to set $\mathbf{x}_{\mathrm{adv}}=f(\mathbf{x},\mathbf{\varepsilon})$ as a hard constraint, so that $\mathbf{x}_\mathrm{adv}$ can be strictly reconstructed. However in practice, this causes difficulty in optimization. Therefore, we relax this constraint and assume $\mathbf{x}_{\mathrm{adv}}\sim \mathcal{N}(f(\mathbf{x},\mathbf{\varepsilon}),\sigma^2\mathbf{I})$, so that
\begin{eqnarray}\label{eqn:reconstruct}
\log p(\mathbf{x}_\mathrm{adv}|\mathbf{x},\mathbf{\varepsilon})=-\frac{1}{2\sigma^2}\|\mathbf{x}_\mathrm{adv}-f(\mathbf{x},\mathbf{\varepsilon})\|_2^2+C_\sigma.
\end{eqnarray}
where $C_\sigma=-\frac{3hw}{2}\log 2\pi\sigma^2$ is irrelevant to the optimization, and can be dropped. There may be some cases where the transformation $f$ has no clear expression, or is non-differentiable w.r.t. $\mathbf{x}$ and/or $\mathbf{\varepsilon}$, such that Eqn.~\ref{eqn:reconstruct} cannot be optimized directly. In such cases, an \textit{auxiliary model} $f_\mathbf{\psi}$ can be built as a substitute, by supervisely training a neural network with tuples of $(\mathbf{x}_\mathrm{adv},\mathbf{x},\mathbf{\varepsilon})$ as training data.

To conclude, the final loss is the combination of above 3 terms:
\begin{eqnarray}\label{eq:all_loss}
\begin{aligned}
&\underset{\mathbf{x},\mathbf{\varepsilon}}{\mathrm{min}}\ \mathcal{L}(\mathbf{x},\mathbf{\varepsilon};\mathbf{x}_\mathrm{adv}) \\
=\ &\mathbb{E}_{\sigma,\mathbf{n}}\ \left\|D_{\mathbf{\theta}}(\mathbf{x}_\sigma; \sigma)-\mathbf{x}\right\|_{2}^{2} \\
+\ &\lambda_1\cdot \phi(\mathbf{\varepsilon}) 
+\lambda_2\cdot \|\mathbf{x}_\mathrm{adv}-f(\mathbf{x},\mathbf{\varepsilon})\|_2^2.
\end{aligned}
\end{eqnarray}
$\lambda_1$ and $\lambda_2$ are hyper-parameters to be tuned, which represents the relative weights. We then discuss the whole NAPPure algorithm for purification.

\subsection{Algorithm}\label{sec:algorithm}

\begin{algorithm}[t]
\caption{Non-additive Perturbation Purification (NAPPure)}
\label{alg:nappure}
\begin{algorithmic}[1]
\REQUIRE Input image $\mathbf{x}_{\mathrm{adv}}$, pretrained EDM model $D_\mathbf{\theta}$, potential function $\phi$, transformation function (or auxiliary model) $f$, identity element $\mathbf{\varepsilon}_0$, weights $\lambda_1,\lambda_2$, number of iterations $T$, learning rates $\eta_1,\eta_2$.
\ENSURE Purified image $\mathbf{x}^*$, perturbation parameter $\mathbf{\varepsilon}^*$
\STATE Initialize $\mathbf{x}^{(0)}\leftarrow\mathbf{x}_{\mathrm{adv}},\mathbf{\varepsilon}^{(0)}\leftarrow\mathbf{\varepsilon}_0$
\FOR {$t=0$ to $T-1$}
\STATE Sample $\mathbf{n}\sim \mathcal{N}(\mathbf{0},\mathbf{I}), \sigma\sim\mathcal{U}(0.4, 0.6)$
\STATE $\mathbf{x}^{(t)}_\sigma\leftarrow \mathbf{x}^{(t)}+\sigma\cdot\mathbf{n}$
\STATE $\mathcal{L}(\mathbf{x}^{(t)};\mathbf{x}_\mathrm{adv},\mathbf{\varepsilon}^{(t)},\mathbf{n},\sigma)\leftarrow\left\|D_{\mathbf{\theta}}(\mathbf{x}^{(t)}_\sigma; \sigma)-\mathbf{x}^{(t)}\right\|_{2}^{2}+\lambda_2\cdot \|\mathbf{x}_\mathrm{adv}-f(\mathbf{x}^{(t)},\mathbf{\varepsilon}^{(t)})\|_2^2$
\STATE $\Delta\mathbf{x}^{(t)}\leftarrow\nabla_\mathbf{x}\ \mathcal{L}(\mathbf{x}^{(t)};\mathbf{x}_\mathrm{adv},\mathbf{\varepsilon}^{(t)},\mathbf{n},\sigma)$ 
\STATE Update image $\mathbf{x}^{(t+1)}\leftarrow \mathrm{Adam}(\mathbf{x}^{(t)},\Delta\mathbf{x}^{(t)},\eta_1)$
\STATE $\mathcal{L}(\mathbf{\varepsilon}^{(t)};\mathbf{x}_\mathrm{adv},\mathbf{x}^{(t)})\leftarrow\lambda_1\cdot \phi(\mathbf{\varepsilon}^{(t)})+\lambda_2\cdot \|\mathbf{x}_\mathrm{adv}-f(\mathbf{x}^{(t)},\mathbf{\varepsilon}^{(t)})\|_2^2$
\STATE $\Delta\mathbf{\varepsilon}^{(t)}\leftarrow\nabla_\mathbf{\varepsilon}\ \mathcal{L}(\mathbf{\varepsilon}^{(t)};\mathbf{x}_\mathrm{adv},\mathbf{x}^{(t)})$ 
\STATE Update parameter $\mathbf{\varepsilon}^{(t+1)}\leftarrow \mathrm{Adam}(\mathbf{\varepsilon}^{(t)},\Delta\mathbf{\varepsilon}^{(t)},\eta_2)$
\ENDFOR
\RETURN $\mathbf{x}^{(T)}, \mathbf{\varepsilon}^{(T)}$
\end{algorithmic}
\end{algorithm}

Given an input image $\mathbf{x}_{\mathrm{adv}}$, we initialize $\mathbf{x}$ the same as $\mathbf{x}_{\mathrm{adv}}$, and initialize $\mathbf{\varepsilon}$ to the identity element of transformation $\mathbf{\varepsilon}_0$. Then we update $\mathbf{x}$ and $\mathbf{\varepsilon}$ alternately according to the loss in Eqn.~\ref{eq:all_loss} using Adam optimizer \cite{kingma2014adam} till convergence. We sample one $\mathbf{n}$ and $\sigma$ instead of calculating the expectation in Eqn.~\ref{eq:all_loss} for efficiency. The final $\mathbf{x}$ is regarded as purified image and can be fed into the downstream classifier. The main procedure is shown in Fig.~\ref{fig:main}, and details can be found in Alg.~\ref{alg:nappure}.

We implement four versions of NAPPure algorithm corresponding to different perturbation types: one for additive transformation, three for each non-additive transformations introduced in Sec.~\ref{sec:setting}. Below are details of each of them.
\begin{itemize}[leftmargin=*, align=left]
\item \textbf{Additive.} This is the version for standard adversarial attack, where $\mathbf{\varepsilon}\in\mathbb{R}^{3hw}$ is the additive noise. We set the potential function to squared $l_2$ norm $\phi(\mathbf{\varepsilon})=\|\mathbf{\varepsilon}\|^2_2$, and the identity element to $\mathbf{\varepsilon}_0=\mathbf{0}$.
\item \textbf{Convolution.} In this version, $\phi(\mathbf{\varepsilon})$ is the convolution kernel. We set the identity element $\mathbf{\varepsilon}_0$ to a kernel with one $1$ in the kernel center and $0$ elsewhere. The potential function is then set to squared $l_2$ distance to the identity element $\phi(\mathbf{\varepsilon})=\|\mathbf{\varepsilon}-\mathbf{\varepsilon}_0\|^2_2$.
\item \textbf{Patch.} In this version, $\mathbf{\varepsilon}=(\mathbf{p},a,b,s)$ is a combination of patch pattern $\mathbf{p}$, coordinates $a,b$, and patch size $s$. We set the identity element to $\mathbf{\varepsilon}_0=(\mathbf{x}_{\mathrm{adv}}, \frac{h}{2}, \frac{w}{2}, 0)$, and the potential function to $\phi(\mathbf{\varepsilon})=|s|$. Since the transformation $f$ is generally non-differentiable w.r.t. $a,b,s$, we train an auxiliary model to approximate $f$, while the training data is crafted with clean images and random parameters.
\item \textbf{Flow-field.} In this version, $\mathbf{\varepsilon}$ is a 2-dimensional flow field. We also set the potential function to squared $l_2$ norm $\phi(\mathbf{\varepsilon})=\|\mathbf{\varepsilon}\|^2_2$, and the identity element to $\mathbf{\varepsilon}_0=\mathbf{0}$.
\end{itemize}

Note that the above versions are illustrative examples to show how to configure our NAPPure framework for specific transformations. NAPPure is not limited to such implementation and can be used for general types of perturbations, as long as the transformation function, potential function, and identity element are provided.

\begin{figure*}[t]
    \centering
    \includegraphics[width=1.8\columnwidth]{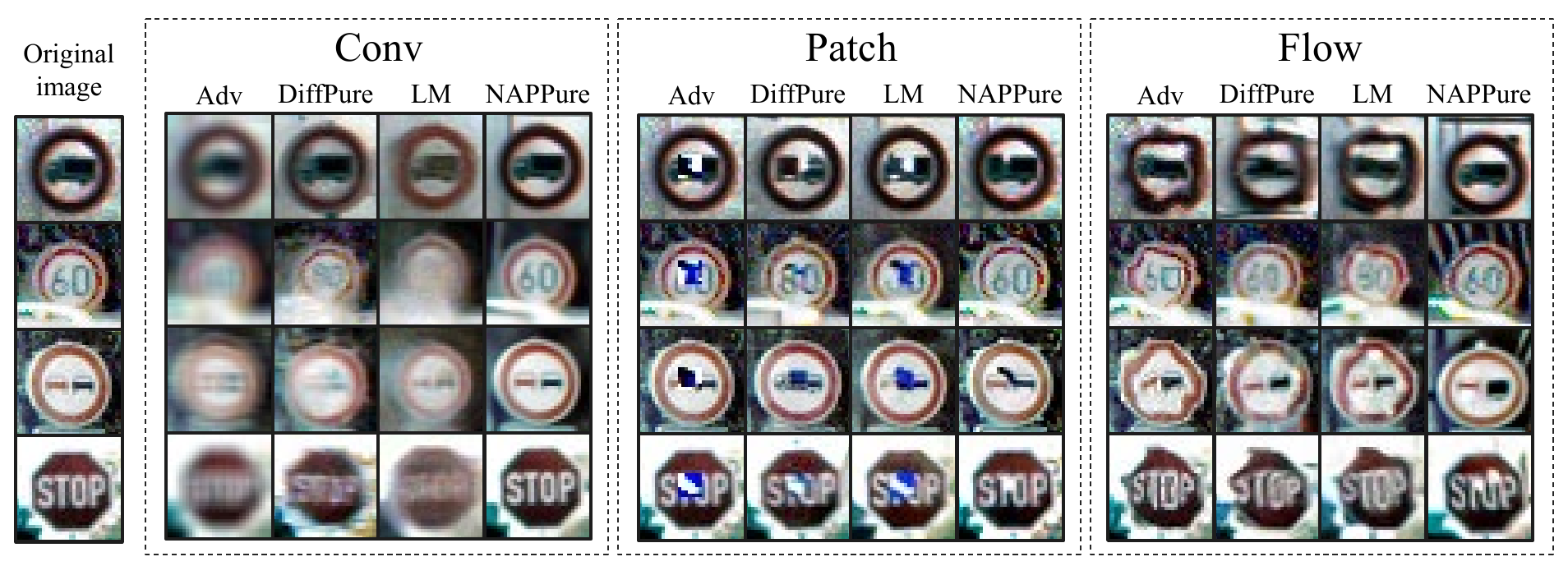}
    \caption{Purification results of different methods under 3 types of non-additive perturbations. The first column (Adv) in each block are adversarial images w.r.t. the classifier.}
    \label{fig:clean_case}
\end{figure*}

\textbf{Composite transformation.} For complex transformation that is constructed by composition of simple transformations, a straightforward solution is to regard such transformation as a single $f$ during purification. Though effective in our observation, such direct implementation causes difficulty in hyper-parameter tuning due to interference among each perturbations. Instead, we suggest a substitute solution for better stability, which replace each simple transformation by a interpolation between perturbed image and original image. 
Given a basis transformations $f$ with corresponding parameter $\mathbf{\varepsilon}$, we introduce a learnable weight $w\in[0,1]$, and construct the new transformation as 
\begin{equation}
\hat{f}(\mathbf{x},\hat{\mathbf{\varepsilon}})=w\cdot f(\mathbf{x},\mathbf{\varepsilon}) + (1-w)\cdot \mathbf{x},
\end{equation}
where $\hat{\mathbf{\varepsilon}}=[w,\mathbf{\varepsilon}]$. Then the composite transformation is constructed by $f=\hat{f}_n\circ\cdots\circ\hat{f}_1$ with $\mathbf{\varepsilon}=[w_1,\cdots,w_n, \mathbf{\varepsilon}_1,\cdots,\mathbf{\varepsilon}_n]$. We name this method \textit{NAPPure-joint}, to distinguish it from original NAPPure algorithm that simply composite all transformations directly.

\textbf{Degeneration of the \textit{Additive} case.} We review our NAPPure algorithm when the transformation function is additive, i.e., $f(\mathbf{x},\mathbf{\varepsilon})=\mathbf{x}+\mathbf{\varepsilon}$. Rewrite Eqn.~\ref{eqn:objective} into two layers of optimization and expand only the last term, there is 
\begin{equation}
\underset{\mathbf{x}}{\text{max}}\ \underset{\mathbf{\varepsilon}}{\text{max}} \log p(\mathbf{x}) +\log p(\mathbf{\varepsilon}) -\lambda_2\cdot \|\mathbf{x}_\mathrm{adv}-\mathbf{x}-\mathbf{\varepsilon}\|_2^2.
\end{equation}
Further assume a uniform prior $p(\mathbf{\varepsilon})\equiv \frac{1}{\mu(\Omega)}$ where $\mu(\cdot)$ is the Lebesgue measure, and the parameter domain $\Omega$ is large enough to cover any possible $\mathbf{\varepsilon}$, then the inner optimization problem 
\begin{equation}
\underset{\mathbf{\varepsilon}}{\text{max}} \log p(\mathbf{\varepsilon}) -\lambda_2\cdot \|\mathbf{x}_\mathrm{adv}-\mathbf{x}-\mathbf{\varepsilon}\|_2^2
\end{equation}
takes its maximum value $-\log \mu(\Omega)$ with $\varepsilon=\mathbf{x}-\mathbf{x}_\mathrm{adv}$. In this case, the original optimization problem degenerates into a simple form, i.e. $\underset{\mathbf{x}}{\text{max}}\log p(\mathbf{x})$, which is exactly the purification objective of LM \cite{chen2023robust}, one of the standard adversarial purification methods. This result indicates that previous adversarial purification methods are inherently designed for additive perturbations, which is one specific case in our NAPPure framework. 
\section{Experiment}

\begin{table*}[t]
\centering
\begin{tabular}{l|cc|cc|cc|cc|cc}
\toprule[1.5pt]
Defense& \multicolumn{2}{c|}{Conv} & \multicolumn{2}{c|}{Patch} & \multicolumn{2}{c|}{Flow} & \multicolumn{2}{c|}{Add}& \multicolumn{2}{c}{Avg}\\
Method&Acc & Rob& Acc & Rob & Acc & Rob & Acc & Rob & Acc & Rob\\
\midrule[1pt]
None \cite{eykholt2018robust} & 95.11&57.42&95.11&13.67&95.11&1.56&95.11&3.12&95.11&18.94\\
\midrule
AT \cite{mkadry2017towards} & 89.45&\underline{61.72}&92.57&19.92&91.60&19.72&88.28&47.85&90.48&37.30\\
DiffPure\textsuperscript{*}  \cite{nie2022diffusion} & 89.26&61.52&89.26&\underline{46.29}&89.26&\underline{21.88}&89.26&60.74&89.26&\underline{47.61}\\
LM\textsuperscript{*}  \cite{chen2023robust} & 93.16&53.32&93.16&13.67&93.16&8.79&93.16&\underline{79.07}&93.16&38.71\\
\midrule
NAPPure & 94.53&\textbf{86.91}&93.55&\textbf{74.22}&93.36&\textbf{51.37}&93.75&\textbf{83.20}&93.80&\textbf{73.93}\\
NAPPure-joint\textsuperscript{*}  & 93.75&76.17&93.75&57.23&93.75&37.37&93.75&66.40&93.75&59.29\\
\bottomrule[1.5pt]
\end{tabular}
\caption{Clean accuracy (Acc \%) and robust accuracy (Rob \%) of different methods against adversarial attacks with different types of perturbations on GTSRB dataset. Methods marked with * share identical implementation across attack types.}
\label{gtsrbmain}
\end{table*}

\begin{table*}[t]
\centering
\begin{tabular}{l|cc|cc|cc|cc|cc}
\toprule[1.5pt]
Defense& \multicolumn{2}{c|}{Conv} & \multicolumn{2}{c|}{Patch} & \multicolumn{2}{c|}{Flow} & \multicolumn{2}{c|}{Add}& \multicolumn{2}{c}{Avg}\\
Method&Acc & Rob& Acc & Rob & Acc & Rob & Acc & Rob & Acc & Rob\\
\midrule[1pt]
None \cite{zagoruyko2016wide} & 96.68&9.57&96.68&10.74&96.68&0.00&96.68&0.00&96.68&5.08\\
\midrule
AT \cite{mkadry2017towards} &78.32&20.11&87.30&\underline{76.37}&74.22&14.25&79.88&35.35&79.93&36.52\\
DiffPure\textsuperscript{*}  \cite{nie2022diffusion} & 89.26&59.38&89.26&69.73&89.26&\underline{23.06}&89.26&\underline{79.10}&89.26&\underline{57.82}\\
LM\textsuperscript{*}  \cite{chen2023robust} &84.96&\underline{60.16}&84.96&36.13&84.96&13.09&84.96&70.12&84.96&44.88\\
\midrule
NAPPure & 91.92&\textbf{66.40}&90.42&\textbf{76.75}&84.38&\textbf{48.24}&89.65&\textbf{82.81}&89.09&\textbf{68.55}\\
NAPPure-joint\textsuperscript{*}  & 87.30&60.54&87.30&76.37&87.30&25.39&87.30&76.56&87.30&59.72\\
\bottomrule[1.5pt]
\end{tabular}
\caption{Clean accuracy (Acc \%) and robust accuracy (Rob \%) of different methods against adversarial attacks with different types of perturbations on CIFAR-10 dataset. Methods marked with * share identical implementation across attack types.}
\label{cifar10main}
\end{table*}

\subsection{Settings}

\textbf{Datasets and models.} We perform our experiments on two real-world image classification datasets: GTSRB dataset \cite{stallkamp2012man} which contains 43 types of traffic signs, and CIFAR-10 dataset \cite{krizhevsky2009learning} which contains 10 types of general objects. All images are of size 32x32 with RGB channels. For each dataset, we randomly select 512 images from test set for evaluation. For the downstream classifier, we use off-the-shelf pre-trained models, including GTSRB-CNN \cite{eykholt2018robust} for GTSRB dataset, and WideResNet \cite{zagoruyko2016wide} for CIFAR-10 dataset. For diffusion models in adversarial purification methods, we use the same EDM architecture \cite{karras2022elucidating} and train them on each dataset following the training scheme in the work of \citet{chen2023robust}.

\textbf{Baseline methods.} We compare our NAPPure algorithm with several strong adversarial defense methods:
\begin{itemize}
    \item AT \cite{mkadry2017towards}: A standard adversarial training method that takes constructed adversarial examples as training data.
    \item DiffPure \cite{nie2022diffusion}: An adversarial purification method that utilizes the backward process of diffusion models. We replace their score-SDE backbone by EDM, for both improved performance and fair comparison.
    \item LM \cite{chen2023robust}: An adversarial purification method that directly maximizes the log-likelihood. We discard the diffusion classifier and only use their purification algorithm for fair comparison.
\end{itemize}
All methods are assumed to know the perturbation types in advance, including: AT constructed adversarial examples with corresponding perturbation type; our NAPPure uses the corresponding version for purification. We also report the results of the type-agnostic version of our NAPPure (NAPPure-joint), which composites all 4 types of transformations according to the technique in Sec.~\ref{sec:algorithm}.

\textbf{Adversarial attacks.} We apply adversarial attacks with the aforementioned types of perturbations in Sec.~\ref{sec:setting}, with the following configurations:

\begin{itemize}
    \item Conv: A convolution-based blur attack using a 5×5 uniform kernel $\mathbf{\varepsilon}_0$, with attack parameters constrained by $\|\mathbf{\varepsilon}-\mathbf{\varepsilon}_0\|_\infty\leq 0.025$.
    \item Patch: The patch based occlusion attack. The patch is fixed at the center of the image with a fixed size 7x7.
    \item Flow: The flow-field based distortion attack. Parameters are limited by $\|\mathbf{\varepsilon}\|_\infty\leq 1.2$ for GTSRB and $\|\mathbf{\varepsilon}\|_\infty\leq 3$ for CIFAR-10. To ensure natural-looking of the distortion, we apply Gaussian smoothing with standard deviation $1.5$ onto the parameters, before the flow-field transformation. The kernel size is 9x9 for GTSRB and 5x5 for CIFAR-10.
    \item Add: The traditional adversarial attack with additive perturbations. Parameters are limited by $\|\mathbf{\varepsilon}\|_\infty\leq 24/255$ for GTSRB, and $\|\mathbf{\varepsilon}\|_\infty\leq 8/255$ for CIFAR-10.
\end{itemize}
We use the widely-adopted adversarial attack method APGD-CE \cite{croce2020reliable}, and follow the objective in Sec.~\ref{sec:setting} to generate white-box adversarial examples for each defense method.

\textbf{Hyper-parameters.} For our NAPPure algorithm, we set learning rate to $\eta_1=0.1, \eta_2=0.05$, and purification steps to $T=500$ in all experiments. For the weights $\lambda_1$ and  $\lambda_2$, we run grid-search in each experiment, and select the pair with highest accuracy on adversarial examples. Such adversarial examples are constructed by performing transfer attack on the raw classifier using 512 images from validation set. See Sec.~\ref{sec:analysis} for detailed results.

\subsection{Main Results}
We report the classification accuracy on clean images (clean accuracy) and that under adversarial attacks (robust accuracy) in Tab.~\ref{gtsrbmain} and Tab.~\ref{cifar10main}. Our main observations are as follows:
\begin{itemize}
    \item Standard adversarial purification methods are much less effective under non-additive attacks. While DiffPure and LM achieve high robustness under additive attacks, they achieve much lower value under non-additive settings. 
    \item Our NAPPure algorithm is effective against all attack types (73.93\% average robust accuracy on GTSRB), especially for non-additive ones. NAPPure achieves comparable robust accuracy to DiffPure and LM under additive attacks, and much higher robust accuracy under the Conv/Patch/Flow settings ($>$25\% average boost on GTSRB). Meanwhile, NAPPure outperforms AT under nearly all settings in terms of robust accuracy, showing its superiority ($>$35\% average boost on GTSRB).
    \item NAPPure is also effective when the exact perturbation type is unknown but falls into a set of known types. This is indicated by the robust accuracy of NAPPure-joint, whose performance is inferior to NAPPure due to missing information, but is still superior to other baseline methods under most attacks ($>$10\% average boost on GTSRB).
\end{itemize}
Above conclusions are more apparent on GTSRB dataset than CIFAR-10, this is reasonable since classifying traffic signs rely more on clear shape boundaries, which is more sensitive to non-additive perturbations. Such results demonstrate the superiority of our NAPPure algorithm under adversarial attacks with non-additive perturbations. 

\subsection{Analytical Results}\label{sec:analysis}

\textbf{Purification results.} We show some examples of the purified images under different types of perturbations on GTSRB dataset in Fig. \ref{fig:clean_case}. NAPPure successfully recovers semantic details from non-additively perturbed images. For blurred inputs, NAPPure recovers sharp edges and fine textures, e.g., boundaries of traffic sign digits become clearer and recognizable. This contrasts with standard adversarial purification methods, which often fails to remove the blur effect. Similarly, for occluded images, NAPPure successfully inpaints adversarial patches, revealing underlying content such as vehicle shapes in GTSRB, while DiffPure and LM fail catastrophically. Geometric distortions are also corrected with our NAPPure: deformed objects recover canonical shapes and spatial relationships (e.g., realigned traffic sign symmetry), while baseline methods either retain the distortion or significantly change the semantics. These visual illustrations align with the improvements in our quantitative results, demonstrating NAPPure’s ability to disentangle perturbations through joint optimization of clean images and transformation parameters, avoiding semantic drift and outperforming standard adversarial purification methods in terms of both fidelity and robustness.

\begin{figure}[t]
    \centering
    \includegraphics[width=0.8\columnwidth]{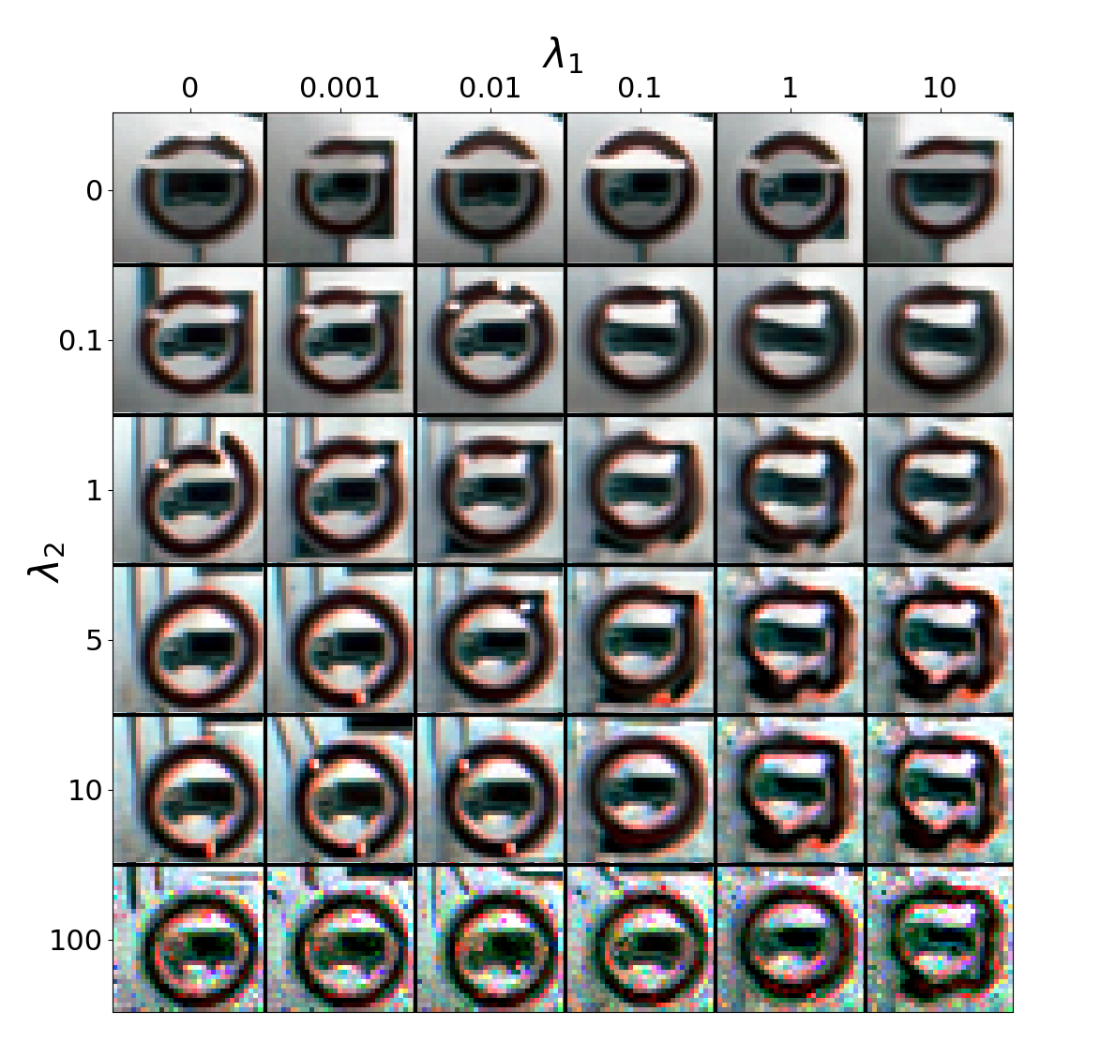}
    \caption{Purification results for different hyper-parameters.}
    \label{flow_lambda}
\end{figure}

\begin{figure}[t]
\vskip -0.1in
    \centering
    \includegraphics[width=0.9\columnwidth]{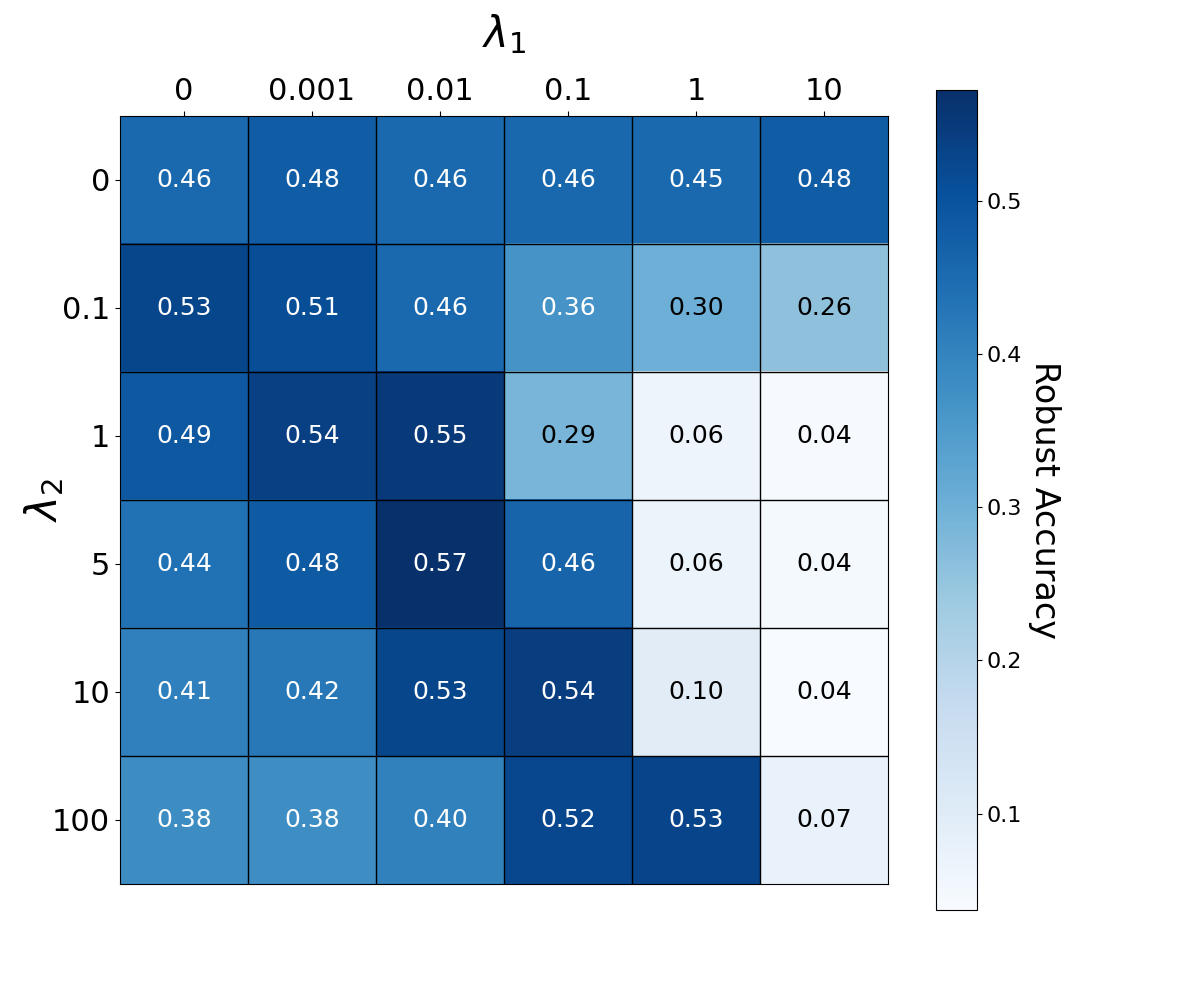}
    \caption{Classification accuracy for different hyper-parameters. }
    \label{flow_lambda_robust_acc}
\vskip -0.1in
\end{figure}

\textbf{Effect of $\lambda$.} We take the flow-field based attacks on GTSRB dataset as an illustrative example. Fig.~\ref{flow_lambda} and Fig.~\ref{flow_lambda_robust_acc} demonstrate the influences of hyper-parameters $\lambda_1$ and $\lambda_2$ in terms of purification quality and robust accuracy.
Lower values of $\lambda_1$ ($\lambda_1 < 0.01$) cause the purification to develop in an unconstrained manner, i.e. while combating distortion in the input image, new distortions are introduced. In contrast, higher values of $\lambda_1$ ($\lambda_1 > 0.1$) overly constrain the perturbation magnitude, resulting in incomplete purification with residual distortions. When $\lambda_2$ is low ($\lambda_2 < 1$), the reconstruction term fails to function effectively, leading to uncertain purification direction, which causes over-smoothing and semantic degradation. Conversely, extremely high values of $\lambda_2$ ($\lambda_2 > 10$) tightly couple the reconstruction term with the adversarial input, limiting the flexibility of purification and introducing new noises in the image.
The interaction between $\lambda_1$ and $\lambda_2$ is of vital significance. For flow-field attacks on GTSRB dataset, the combination $\lambda_1 = 0.01$ and $\lambda_2 = 5$ achieves the optimal balance. $\lambda_1$ adequately regularizes the magnitude of flow-field without over-constraint, while $\lambda_2$ ensures the geometric fidelity of the purified images.

\begin{table}[t]
\centering
\begin{tabular}{l|c}
\toprule[1.5pt]
Defense Method & Robust Accuracy\\
\midrule[1pt]
None \cite{eykholt2018robust} & 12.70\\
DiffPure \cite{nie2022diffusion} & 30.00\\
LM \cite{chen2023robust} & 15.82\\
\midrule
NAPPure & 37.10\\
NAPPure-joint & \textbf{54.49}\\
\bottomrule[1.5pt]
\end{tabular}
\caption{Robust accuracy (\%) against adversarial attacks with composition of all 4 types of perturbations on GTSRB dataset.}
\label{tab:composite}
\end{table}

\textbf{Composite Attacks.} We also examine the effectiveness of the NAPPure-joint algorithm against adversarial attacks with composite transformations. We apply all 4 types of perturbations in a single attack, while the magnitude of each single perturbation is reduced to avoid severe semantic shift. 
The results are shown in Tab.~\ref{tab:composite}. While baseline methods exhibits clear weakness under such attacks, NAPPure-joint still achieves certain degree of robustness, exhibiting the ability to defend against composite attacks. NAPPure-joint also outperforms NAPPure, indicating the effectiveness of our interpolation technique.

\section{Conclusion}

In this paper, we proposed the NAPPure framework to address the challenge of adversarial attacks under non-additive perturbations in image classification. By modeling the generation process of perturbed images and disentangling clean images and perturbation parameters through likelihood maximization, NAPPure achieves remarkable effectiveness in enhancing model robustness against various non-additive perturbations. Meanwhile, NAPPure exactly degenerates into traditional adversarial purification method under additive perturbations, contributing a compatible extension to existing approaches.

\section*{ACKNOWLEDGEMENT}
This work was supported by the Strategic Priority Research Program of the Chinese Academy of Sciences (Grant No. XDB0680101), the Youth Program of the National Natural Science Foundation of China (Grant No. 62406309), and CAS Project for Young Scientists in Basic Research (Grant No. YSBR-034).

{
    \small
    \bibliographystyle{ieeenat_fullname}
    \bibliography{main}
}

\newpage 

\section*{Appendix}

\subsection*{A1. Additional Experimental Results on ImageNet}

To further verify the scalability of NAPPure, we conducted a large-scale experiment on ImageNet dataset. We sample 512 samples for evaluation. For the diffusion model in the adversarial purification method, we adopted the pre-trained unconditional diffusion model provided by Karras et al. (2022) \cite{karras2022elucidating}, and for the classifier, we followed the ResNet-50 framework used by Nie et al. (2022) \cite{nie2022diffusion}. As shown in Tab.~\ref{ImgNet}, we performed experiments using four types of attacks, with the detailed configurations of these attack types as follows:
\begin{itemize}
    \item Conv: A convolution-based blur attack using a 15×15 uniform kernel $\mathbf{\varepsilon}_0$, with attack parameters constrained by $\|\mathbf{\varepsilon}-\mathbf{\varepsilon}_0\|_\infty\leq 0.025$.
    \item Patch: The patch-based occlusion attack. The patch is fixed at the center of the image with a fixed size 50x50.
    \item Flow: The flow-field based distortion attack. Parameters are limited by $\|\mathbf{\varepsilon}\|_\infty\leq 1.2$ . To ensure natural-looking of the distortion, we apply Gaussian smoothing with standard deviation $1.5$ onto the parameters, before the flow-field transformation. The kernel size is 29x29.
    \item Add: The traditional adversarial attack with additive perturbations. Parameters are limited by $\|\mathbf{\varepsilon}\|_\infty\leq 4/255$ .
\end{itemize}

\textbf{Result}. As shown in Table \ref{ImgNet}, our NAPPure method outperforms DiffPure method by 8.19\%. This indicates that our method is also effective on large-scale datasets.

\subsection*{A2. More details of the experiments}
Table \ref{tab:parameter_settings} summarizes the detailed parameter settings used in our evaluations on the GTSRB and CIFAR-10 datasets. 

Specifically, this table outlines the number of iterations, as well as the values of regularization parameters \(\lambda_1\) (controlling the perturbation prior loss) and \(\lambda_2\) (governing the image reconstruction loss), for each combination of dataset, attack type (Additive, Blur, Flow, Patch), and defense method (NAPPure and NAPPure-joint). The variations in settings across different scenarios (e.g., fewer iterations for Additive attacks on CIFAR-10 compared to non-additive attacks) reflect the need to adapt to the distinct characteristics of each perturbation type and dataset.

\subsection*{A3. Computational Cost Analysis}

Purification efficiency holds significant importance for real-world deployment scenarios. To delve into this, we conducted an analysis of the trade-off between the number of purification iterations and model robustness, using the GTSRB dataset under patch attacks as the test case. The detailed results are presented in Table \ref{tab:iterations}.

The findings reveal that NAPPure reaches a near-optimal performance level within 200 iterations, achieving a robust accuracy of 72.26\%. This is merely 1.96\% lower than the 74.22\% robust accuracy obtained after 500 iterations. However, when the number of iterations is extended to 1000, a noticeable performance degradation occurs, with the robust accuracy dropping to 60.74\%. This decline is likely attributed to the over-optimization of perturbation parameters during the extended purification process.

\begin{table}[!h]
    \centering
    \begin{tabular}{c|c}
    \toprule
    Iterations & Robust Acc \\
    \midrule
    100 & 63.48\% \\
    200 & 72.26\% \\
    500 & 74.22\% \\
    1000 & 60.74\% \\
    \bottomrule
    \end{tabular}
    \caption{The robust accuracy under different numbers of purification iterations (GTSRB, patch attack).}
    \label{tab:iterations}
\end{table}

\begin{table}[!h]
    \centering
    \begin{tabular}{lcc}
    \toprule
    Auxiliary Model & Robust Acc & Clean Acc \\
    \midrule
    3-layer CNN & 74.22\% & 93.55\% \\
    ResNet-18 & 71.29\% & 93.16\% \\
    \bottomrule
    \end{tabular}
    \caption{Impact of auxiliary model architecture on NAPPure performance (GTSRB, patch attack).}
    \label{tab:aux}
\end{table}

\begin{table}[!h]
    \centering
    \begin{tabular}{lcc}
    \toprule
    Attack Type & Attack Parameter & Robust Acc \\
    \midrule
    \centering Patch Attack & 5×5 & 85.16\% \\
     & 7×7 & 74.22\% \\
     & 9×9 & 67.97\% \\
    \midrule
    \centering Blur Attack & 3×3 & 91.80\% \\
     & 5×5 & 86.91\% \\
    \bottomrule
    \end{tabular}
    \caption{Generalization of NAPPure to varying attack parameters (GTSRB).}
    \label{tab:conv_params}
\end{table}
\begin{table*}[!ht]
\centering
\begin{tabular}{l|cc|cc|cc|cc|cc}
\toprule[1.5pt]
Defense& \multicolumn{2}{c|}{Conv} & \multicolumn{2}{c|}{Patch} & \multicolumn{2}{c|}{Flow} & \multicolumn{2}{c|}{Add}& \multicolumn{2}{c}{Avg}\\
Method&Acc & Rob& Acc & Rob & Acc & Rob & Acc & Rob & Acc & Rob\\
\midrule[1pt]

None & 75.78&11.33&75.78&7.81&75.78&0&75.78&0&75.78&4.79\\

\midrule
DiffPure\textsuperscript{*}  \cite{nie2022diffusion} & 69.92&20.83&69.92&42.97&69.92&7.81&69.92&46.88&69.92&29.62\\

LM\textsuperscript{*}  \cite{chen2023robust} & 67.97&12.11&67.97&6.25&67.97&17.97&67.97&59.38&67.97&23.93\\

\midrule
NAPPure & 69.11&\textbf{21.48}&65.26&\textbf{48.05}&68.35&\textbf{21.48}&69.33&\textbf{60.16}&68.01&\textbf{37.79}\\

\bottomrule[1.5pt]
\end{tabular}
\caption{Clean accuracy (Acc \%) and robust accuracy (Rob \%) of different methods against adversarial attacks with different types of perturbations on ImgNet dataset. Methods marked with * share identical implementation across attack types.}
\label{ImgNet}
\end{table*}

\begin{table*}[!ht]
    \centering
    \begin{tabular}{c|c|c|c|c|c}
    \toprule
    \textbf{Dataset} & \textbf{Attack Type} & \textbf{Defense Method} & \textbf{Iterations} & \boldmath{$\lambda_1$} & \boldmath{$\lambda_2$} \\
    \midrule
    GTSRB & Additive & NAPPure & 100 & 0.1 & 3 \\
    GTSRB & Blur & NAPPure & 500 & 0.001 & 3 \\
    GTSRB & Flow & NAPPure & 500 & 0.01 & 1 \\
    GTSRB & Patch & NAPPure & 500 & 0.01 & 5 \\
    GTSRB & - & NAPPure-joint & 500 & 0.001 & 3 \\
    \midrule
    CIFAR-10& Additive & NAPPure & 20 & 0.1 & 5 \\
    CIFAR-10& Blur & NAPPure & 500 & 0.001 & 5 \\
    CIFAR-10& Flow & NAPPure & 500 & 0.01 & 1 \\
    CIFAR-10& Patch & NAPPure & 500 & 0.01 & 5 \\
    CIFAR-10& - & NAPPure-joint & 100 & 0.01 & 5 \\
    \midrule
    ImageNet& Additive & NAPPure & 10 & 0.1 & 3 \\
    ImageNet& Blur & NAPPure & 100 & 0.01 & 5 \\
    ImageNet& Flow & NAPPure & 100 & 0.01 & 1 \\
    ImageNet& Patch & NAPPure & 100 & 0.01 & 10 \\
    \bottomrule
    \end{tabular}
    \caption{Detailed parameter settings for NAPPure and NAPPure-joint under different attacks on GTSRB and CIFAR-10 datasets}
    \label{tab:parameter_settings}
\end{table*}

\subsection*{A4. The robustness verification of the NAPPure auxiliary model for architectural changes}

The auxiliary model in NAPPure (used for non-differentiable perturbations like patch occlusion) is designed as an image-to-image generative network. To validate its robustness to architectural variations, we compared two architectures: a lightweight 3-layer CNN and a deeper ResNet-18, under patch attacks on GTSRB.

Table \ref{tab:aux} shows that replacing the 3-layer CNN with ResNet-18 results in a minor drop in robust accuracy (74.22\% → 71.29\%, a 2.93\% difference), while clean accuracy remains stable. This insensitivity to architecture arises because the auxiliary model focuses on reconstructing perturbed images rather than discriminative tasks, making it less vulnerable to architectural changes. Importantly, both configurations outperform baselines (e.g., DiffPure’s 46.29\% robust accuracy for patch attacks), confirming the reliability of NAPPure’s design.

\subsection*{A5. The generalization ability of NAPPure under different attack parameters}
A key advantage of NAPPure is its ability to maintain robustness under varying attack parameters, even when the attack parameters differ from those used in defense configuration. We evaluate this generalization capability for two representative non-additive attack types: patch occlusion and convolution-based blur.

For patch attacks, we test NAPPure with a fixed defense model (configured for general patch occlusion) against varying attack patch sizes. NAPPure achieves robust accuracies of 85.16\%, 74.22\%, and 67.97\% for attack patch sizes of 5×5, 7×7, and 9×9, respectively. All results outperform baseline methods (e.g., DiffPure and LM) under the same settings. This is because NAPPure features an adaptive learning mechanism for patch sizes, endowing it with the ability to adapt to different attack scenarios. Such adaptability ensures its effectiveness even when attack patch sizes vary.

For convolution-based blur attacks, we use a defense model with a fixed 5×5 kernel and evaluate against attacks with different kernel sizes. As shown in Table \ref{tab:conv_params}, NAPPure achieves 91.80\% robust accuracy against 3×3 attack kernels and 86.91\% against 5×5 attack kernels. These results confirm that NAPPure remains effective as long as the attack kernel size does not exceed the defense kernel size, validating its generalization to varying convolution parameters.

\end{document}